\documentclass{article}

\usepackage[final]{corl_2018} % Uncomment for the camera-ready ``final'' version
\usepackage{graphicx}
\usepackage{caption} 
\usepackage{algorithm}
\usepackage[noend]{algpseudocode}
\usepackage{amsmath,amssymb}
\usepackage{multirow}
\usepackage{array,multirow,graphicx}
\usepackage{float}
\usepackage{wrapfig}
\usepackage{todonotes}
\usepackage{subcaption}

\newcommand*\rot{\rotatebox{90}}
\newcommand{\policy}{\ensuremath{\pi}}
\newcommand{\expertpolicy}{\ensuremath{\pi^*}}
\newcommand{\action}{\ensuremath{\boldsymbol{a}}}

\newcommand{\obs}{\ensuremath{\boldsymbol{o}}}
\newcommand{\task}{\ensuremath{\mathfrak{T}}}
\newcommand{\example}{\ensuremath{\tau}}

\newcommand{\sentence}{\ensuremath{\boldsymbol{s}}}
\newcommand{\emb}{\ensuremath{f_\theta}}
\newcommand{\loss}{\ensuremath{\mathcal{L}}}
\newcommand{\margin}{\ensuremath{margin}}
\newcommand{\support}{\ensuremath{U}}
\newcommand{\query}{\ensuremath{Q}}
\newcommand{\tecnet}{TecNet}

\title{Task-Embedded Control Networks\\for Few-Shot Imitation Learning}

\author{
  Stephen James \\
  Dyson Robotics Lab \\
  Imperial College London \\
  \texttt{slj12@imperial.ac.uk} \\
  \And
  Michael Bloesch \\
  Dyson Robotics Lab \\
  Imperial College London \\
  \texttt{m.bloesch@imperial.ac.uk} \\
  \And
  Andrew J. Davison \\
  Dyson Robotics Lab \\
  Imperial College London \\
  \texttt{a.davison@imperial.ac.uk} \\
}

\begin{document}
\maketitle

%===============================================================================

\begin{abstract}
Much like humans, robots should have the ability to leverage knowledge from previously learned tasks in order to learn new tasks quickly in new and unfamiliar environments. Despite this, most robot learning approaches have focused on learning a single task, from scratch, with a limited notion of generalisation, and no way of leveraging the knowledge to learn other tasks more efficiently. One possible solution is meta-learning, but many of the related approaches are limited in their ability to scale to a large number of tasks and to learn further tasks without forgetting previously learned ones. With this in mind, we introduce Task-Embedded Control Networks, which employ ideas from metric learning in order to create a task embedding that can be used by a robot to learn new tasks from one or more demonstrations. In the area of visually-guided manipulation, we present simulation results in which we surpass the performance of a state-of-the-art method when using only visual information from each demonstration. Additionally, we demonstrate that our approach can also be used in conjunction with domain randomisation to train our few-shot learning ability in simulation and then deploy in the real world without any additional training. Once deployed, the robot can learn new tasks from a single real-world demonstration.
\end{abstract}


%===============================================================================

\section{Introduction}

\begin{wrapfigure}[22]{r}{0.34\textwidth}
\centering
\includegraphics[width=1.0\linewidth]{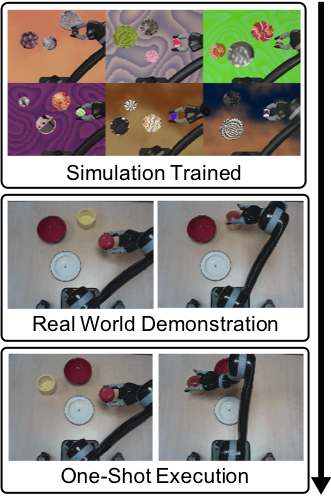}
\caption{The robot gains its few-shot learning ability in simulation, and can then learn a new task from a single demonstration.}
\label{fig:front_summary}
\end{wrapfigure}

Humans and animals are capable of learning new information rapidly from very few examples, and apparently improve their ability to `learn how to learn' throughout their lives~\citep{harlow1949formation}. Endowing robots with a similar ability would allow for a large range of skills to be acquired efficiently, and for existing knowledge to be adapted to new environments and tasks. An emerging trend in robotics is to learn control directly from raw sensor data in an end-to-end manner. Such approaches have the potential to be general enough to learn a wide range of tasks, and they have been shown to be capable of performing tasks that traditional methods in robotics have found difficult, such as when close and complicated coordination is required between vision and control~\citep{levine2016end}, or in tasks with dynamic environments~\citep{james2017transferring}. However, these solutions often learn their skills from scratch and need a large amount of training data~\citep{james2017transferring, zhang2017deep, james20163d}. A significant goal in the community is to develop methods that can reuse past experiences in order to improve the data efficiency of these methods. 

To that end, one significant approach is Meta-Imitation Learning (MIL)~\citep{finn2017one}, in which a policy is learned that can be quickly adapted, via one or few gradient steps at test time, in order to solve a new task given one or more demonstrations. The underlying algorithm, Model-Agnostic Meta-Learning (MAML)~\citep{finn2017model} can be very general, but lacks some of the properties that we might hope for in a robotic system. For one, once the policy is trained, it cannot accomplish any of the tasks seen during training unless it is given an example again at test time. Also, once a specific task is learned, the method can lose its ability to meta-learn and be stuck with a set of weights that can only be used for that one task. One way around this is to make a copy of the weights needed for each task, but this raises scalability concerns. 

Our new approach, Task-Embedded Control Networks (\tecnet s), is centred around the idea that there is an embedding of tasks, where tasks that are similar (in terms of visual appearance) should be close together in space, whilst ones that are different should be far away from one another. Having such an expressive space would not only allow for few-shot learning, but also opens the possibility of inferring information from new and unfamiliar tasks in a zero-shot fashion, such as how similar a new task may be to a previously seen one. 

\begin{figure}
\centering
\includegraphics[width=1.0\linewidth]{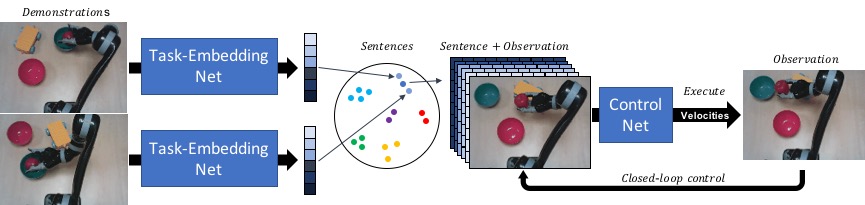}
\caption{Task-Embedded Control Networks (\tecnet s) allow tasks to be learned from single or multiple demonstrations. Images of demonstrations are embedded into a compact representation of a task, which can be combined to create a \textit{sentence}. This sentence is then expanded and concatenated (channel-wise) to the most recent observation from a new configuration of that task before being sent through the control network in a closed-loop manner. Both the task-embedding net and control net are jointly optimised to produce a rich embedding.}
\label{fig:approach}
\end{figure}

\tecnet s, which are summarised in Figure \ref{fig:approach}, are composed of a \textit{task-embedding network} and a \textit{control network} that are jointly trained to output actions (e.g. motor velocities) for a new variation of an unseen task, given a single or multiple demonstrations. The task-embedding network has the responsibility of learning a compact representation of a task, which we call a \textit{sentence}. The control network then takes this (static) sentence along with current observations of the world to output actions. \tecnet s do not have a strict restriction on the number of tasks that can be learned, and do not easily forget previously learned tasks during training, or after. The setup only expects the observations (e.g. visual) from the demonstrator during test time, which makes it very applicable for learning from human demonstrations. 

To evaluate our approach, we present simulation results from two experimental domains proposed in MIL~\citep{finn2017one}, and demonstrate that we can train our meta-learning ability in simulation and then deploy in the real world without any additional training. We believe this to be a desirable property given that large amounts of data are needed to end-to-end solutions. Despite being trained to meta-learn in simulation, the robot can learn new tasks from a single demonstration in the real world.

Our contributions in this work are threefold. We demonstrate the ability to one-shot and few-shot learn visuomotor control through the use of \tecnet s in a selection of visually-guided manipulation tasks. Secondly, we show that \tecnet s are able to achieve higher success rates compared to MIL~\citep{finn2017one} when using only visual information from each demonstration. Finally, we demonstrate the first successful method of a few-shot learning approach trained in simulation and transferred to the real world, which we believe is an important direction for allowing large-scale generalisation.

%===============================================================================

\section{Related Work}

Our work lies at the intersection of imitation learning~\citep{schaal1999imitation, argall2009survey} and meta-learning~\citep{thrun2012learning, lemke2015metalearning}. Imitation learning aims to learn tasks by observing a demonstrator. One focus within imitation learning is \textit{behavioural cloning}, in which the agent learns a mapping from observations to actions given demonstrations, in a supervised learning manner~\citep{pomerleau1989alvinn, ross2011reduction}. Another focus is \textit{inverse reinforcement learning}~\citep{ng2000algorithms}, where an agent attempts to estimate a reward function that describes the given demonstrations~\citep{abbeel2004apprenticeship, finn2016guided}. In our work, we focus on behavioural cloning in the context of learning motor control directly from pixels. A common issue in behavioural cloning is the large amount of data needed to train such systems~\citep{james2017transferring}, as well as the fact that tasks are often learned independently, where learning one task does not accelerate the learning of another. Recently, there has been encouraging work to address this problem~\citep{finn2017one}, and our approach provides a further advance.

One-shot and few-shot learning is the paradigm of learning from a small number of examples at test time, and has been widely studied in the image recognition community~\citep{vinyals2016matching, koch2015siamese, santoro2016meta, ravi2017optimization, triantafillou2017few, snell2017prototypical}. Many one-shot and few-shot learning methods in image recognition are a form of meta-learning, where the algorithms are tested on their ability to learn new tasks, rather than the usual machine learning paradigm of training on a single task and testing on held out examples of that task. Common forms of meta-learning include recurrence~\citep{santoro2016meta}, learning an optimiser~\citep{ravi2017optimization}, and more recently Model Agnostic Meta-Learning (MAML)~\citep{finn2017model}. Many works in metric learning, including ours, can be seen as forms of meta-learning~\citep{vinyals2016matching, snell2017prototypical}, in the sense that they produce embeddings dynamically from new examples during test time; the difference to other more common meta-learning approaches is that the embedding generation is fixed after training.

The success of our new approach comes from learning a metric space, and there has been an abundance of work in metric learning for image classification~\citep{kulis2012metric, bellet2013survey}, from which we will summarise the most relevant. Matching Networks~\citep{vinyals2016matching} use an attention mechanism over a learned embedding space which  produces a weighted nearest neighbour classifier given labelled examples (support set) and unlabelled examples (query set). Prototypical Networks~\citep{snell2017prototypical} are similar, but differ in that they represent each class by the mean of its examples (the prototype) and use a squared Euclidean distance rather than the cosine distance. In the case of one-shot learning, matching networks and prototypical networks become equivalent. Our approach is similar in that our sentence (prototype) is created by averaging over the support set, but differs in the way we couple the learned embedding space with the control network. These metric learning methods have all been developed for image classification, and in the visuomotor control domain of our method, we do not explicitly classify sentences, but instead jointly optimise them with a control network.

Recently, \citet{hausman2018learning} proposed learning a skill embedding space via reinforcement learning that led to speed-ups during training time. Although impressive, that method  does not focus on few-shot learning, and the experiments are run within simulation with low dimensional state spaces. Another piece of work that uses embedding spaces is~\citep{sung2017deep}, where a multimodal embedding is learned for point-clouds, language and trajectories. This work involves pre-training the parts of the network, and also relies on accurate models of the world. Our approach has the benefit that we map directly from images to motor actions and train jointly embedding and control networks, with no pre-training. 

In terms of setup, the closest related work to ours is MIL~\citep{finn2017one}, where they apply MAML~\citep{finn2017model} and behaviour cloning to learn new tasks, end-to-end from one visual demonstration. The underlying algorithm, MAML, learns a set of weights that can be quickly adapted to new tasks. If we were to use this approach to retain information we had previously learnt, we would need to hold copies of weights for each task. In comparison, our method relies on storing a compact sentence for every task we want to remember.

%===============================================================================

\section{Task-Embedded Control Networks}
\label{sec:method}

We now formally summarise the notation for our method. A policy $\policy$ for task $\task$ maps observations $\obs$ to actions $\action$, and we assume to have expert policies $\expertpolicy$ for multiple different tasks.  Corresponding example trajectories consist of a series of observations and actions: $\example = [(\obs_1, \action_1), \ldots, (\obs_T, \action_T)]$ and we define each task to be a set of such examples, $\task = \{ \example_1, \cdots, \example_K \}$. \tecnet s aim to learn a universal policy $\policy(\obs, \sentence)$ that can be modulated by a sentence $\sentence$, where $\sentence$ is a learned description of a task $\task$. The resulting universal policy $\policy(\obs, \sentence)$ should emulate the expert policy $\expertpolicy$ for task $\task$.

\subsection{Task Embedding}

\begin{figure}
\centering
\includegraphics[width=0.85\linewidth]{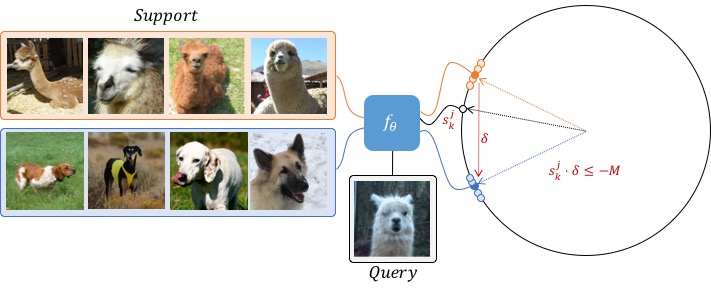}
\caption{A visualisation of how the embedding is learned. Imagine a simple case where we have 2 tasks (or classes): llamas and dogs. We have a support set of 4 examples, which are then embedded and averaged in order to get a sentence for each task. The hinge rank loss drives the dot product of the query image ($\sentence_k^j$) with the difference between the actual sentence and the negative sentences ($\delta$) to be at least a factor of margin away ($M$ in the Figure above). In other words, $\sentence_k^j$ should point in the opposite direction to $\delta$ by at least a factor of margin.}
\label{fig:emb_summary}
\end{figure}

We now introduce our task embedding, which can be used independently in other fields, such as image classification, and so we keep this section general. Assume we are given a small set of $K$ examples of a task $\task^j$. Our task embedding network $\emb: \mathbb{R}^D \rightarrow \mathbb{R}^N$ computes a normalised $N$-dimensional vector $\sentence_k^j \in \mathbb{R}^N$ for each example $\example_k^j \in \task^j$. A combined sentence $\sentence^j \in \mathbb{R}^N$ is then computed for that task by taking the normalised mean of the example vectors:
\begin{align}
\sentence^j = \bigg[ \frac{1}{K} \sum_{\example_k^j \in \task^j} \emb(\example_k^j) \bigg] ^\wedge~,
\label{eq:sentence}
\end{align}
% \label{eq:sentence}
where $\boldsymbol{v}^\wedge = \frac{\boldsymbol{v}}{\|\boldsymbol{v}\|}$. We then need to define a loss function that can be used to learn an ideal embedding. We use a combination of the cosine distance between points and the hinge rank loss (inspired by~\citep{frome2013devise}). The loss for a task $\task^j$ is defined as:
\begin{align}
\loss_{emb} = \sum_{\example_k^j \in \task^j} \sum_{\task^i \neq \task^j} max[0, \margin - \sentence_k^j \cdot \sentence^j + \sentence_k^j \cdot \sentence^i]~,
\end{align}
which trains the model to produce a higher dot-product similarity between a task's example vectors $\sentence_k^j$ and its sentence $\sentence^j$ than to sentences from other tasks $\sentence^i$. We illustrate the intuition behind this loss in Figure \ref{fig:emb_summary}.

Additionally, we pick two disjoint sets of examples for every task $\task^j$: a support set $\task_U^j$ and a query set $\task_Q^j$. In the above embedding loss, the support set is used to compute the task sentences, $\sentence^i$ and $\sentence^j$, and only the examples picked from the query set are used as example vectors, $\sentence_k^j$. Given that each of the sampled tasks in a training batch are unique, the negatives $\task^i$ can be chosen to be all the other tasks in the batch. Therefore, for each task within a batch, we also compare to every other task. Further details are given in Algorithm \ref{alg:training}.

In all of our experiments, we set $\margin = 0.1$, though in practice we found a wide range of values between $0.01 \leq \margin \leq 1.0$ that would work. Although not used, we can treat this embedding as a classification of tasks, whose accuracy we can estimate by computing the sentence $\sentence_k$ of an example and then performing a nearest neighbour search in the embedding space over all task sentences $\sentence^j$. In addition to the dot-product similarity and hinge rank loss, we also tried other distances and losses. One such distance and loss was the squared Euclidean distance used in~\citep{snell2017prototypical}, but we found that this did not work as well for our case.

\subsection{Control}

\begin{algorithm}[t]
\caption{Training loss computation for one batch. $B$ is the batch size, $K_{\support}$ and $K_{\query}$ are the number of examples from the support and query set respectively, and $RandomSample(S, N)$ selects $N$ elements uniformly at random from the set $S$.}
\label{alg:training}
\begin{algorithmic}[1]
\Procedure{Training Iteration}{}
\State $\mathcal{B} = RandomSample(\{\task_1, \cdots, \task_N\}, $B$)$
\For{$\task^j \in \mathcal{B}$}
	\State $\task_\support^j = RandomSample(\task^j, K_{\support})$
    \State $\task_\query^j = RandomSample(\task^j \backslash \task_\support^j, K_{\query})$
    
    \State $\sentence_\support^j = \Big[ \frac{1}{K_{\support}} \sum_{\example \in \task_\support^j} \emb(\example) \Big] ^\wedge$
    \State $\sentence_q^j = \emb(\example_q) \quad \forall \example_q \in \task_\query^j$
\EndFor
$\loss_{emb} = \loss_{ctr}^{\support} = \loss_{ctr}^{\query} = 0$
\For{$\task^j \in \mathcal{B}$}
	\State $\loss_{emb} \mathrel{+}= \sum_{q} \sum_{i \neq j} max[0, \margin - \sentence_q^j \cdot \sentence_\support^j + \sentence_q^j \cdot \sentence_\support^i]$
    \State $\loss_{ctr}^{\support} \mathrel{+}= \sum_{\example \in \task_\support^j} \sum_{(\obs, \action) \in \example} \| \policy(\obs, \sentence_\support^j) - \action \|^2_2$
    \State $\loss_{ctr}^{\query} \mathrel{+}= \sum_{\example \in \task_\query^j} \sum_{(\obs, \action) \in \example} \| \policy(\obs, \sentence_\support^j) - \action \|^2_2$
\EndFor
\State $\loss_{tec} = \lambda_{emb} \loss_{emb} + \lambda_{ctr}^{\support} \loss_{ctr}^{\support} + \lambda_{ctr}^{\query} \loss_{ctr}^{\query}$
\State \textbf{return} $\loss_{tec}$
\EndProcedure
\end{algorithmic}
\end{algorithm}

\begin{algorithm}[t]
\caption{How \tecnet s operate during test time. $D$ is the set of demonstrations for a task, $Env$ is the environment in which to act.}
\label{alg:testing}
\begin{algorithmic}[1]
\Procedure{Test}{$D, Env$}
\State $\sentence = \Big[ \frac{1}{|D|} \sum_{\example \in D} \emb(\example) \Big]^\wedge$
\While{task not complete}
	\State $\obs = Env.GetObservation()$
	\State $\action = \policy(\obs, \sentence)$
    \State $Env.Act(\action)$
\EndWhile
\EndProcedure
\end{algorithmic}
\end{algorithm}

In contrast to metric learning systems for classification, which would use some sort of nearest neighbour test to find the matching class, here the embedding is relayed to the control network and both networks are trained jointly. Given a sentence $\sentence_U^j$, computed from the support set $\task_U^j$, as well as examples from the query set $\task_Q^j$ we can compute the following loss for the policy $\policy$:
\begin{align}
\loss_{ctr} = \sum_{\example_q^j \in \task_Q^j} \sum_{(\obs, \action) \in \example_q^j} \| \policy(\obs, \sentence_U^j) - \action \|^2_2 ~.
\end{align}
This allows the embedding not only to be learned from the embedding loss $\loss_{emb}$, but also from the control loss, which can lead to a more meaningful embedding for the control network than if they were trained independently. Though appropriate weightings must be selected, as the control network needs the embedding in order to know which task to perform, but the embedding network may have to wait for a reasonable control output before being able to enrich its structure.

We found it helpful for the control network to also predict the action for the examples in the support set $\task_\support^j$. This has the advantage that it makes the task of learning $\loss_{ctr}^{\query}$ easier, as learning $\loss_{ctr}^{\support}$ can be seen as an easier version of minimising the former (since example dependent information can be passed through the embedding space). Thus, the final loss is:
\begin{align}
\loss_{Tec} = \sum_{\task} \lambda_{emb} \loss_{emb} + \lambda_{ctr}^{\support} \loss_{ctr}^{\support} + \lambda_{ctr}^{\query} \loss_{ctr}^{\query}
\end{align}

Input to the task-embedding network consists of $(width, height, 3 \times |\example|)$, where $3$ represents the RGB channels. For all of our experiments, we found that we only need to take the first and last frame of an example trajectory $\example$ for computing the task embedding and so discarded intermediate frames, resulting in an input of $(width, height, 6)$. The sentence from the task-embedding network is then tiled and concatenated channel-wise to the input of the control network (as shown in Figure \ref{fig:approach}), resulting in an input image of $(width, height, 3 + $N$)$, where $N$ represents the length of the vector. Pseudocode for both the training and testing is provided in Algorithms \ref{alg:training} and \ref{alg:testing} respectively.
 
%===============================================================================

\section{Experiments}
\label{sec:experiments}

\begin{figure}
\centering
\includegraphics[width=0.85\linewidth]{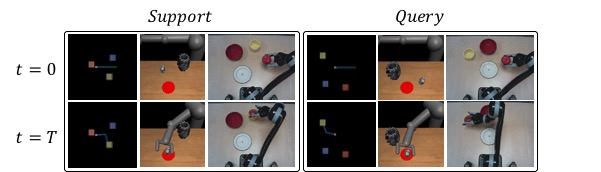}
\caption{Here we show the first and last timestep of a single example (i.e. one-shot) from the support and query set for each of the 3 experimental domains. The support examples are used to describe the task, whilst the query set examples test the networks ability to perform a modified version of the task. We now highlight each of the tasks in the support set. \textit{Left:} the simulated reaching experiment, where the robot must reach a specified colour. \textit{Centre:} the simulated pushing experiment, where the robot must push a specified object to the red target. \textit{Right:} the real world placing experiment, where the robot must place an item into a specified container.}
\label{fig:experiments_summary}
\end{figure}

In this section, we aim to answer the following: (1) Is it possible to learn a task embedding that can directly be used for visuomotor control? (2) Does our metric loss lead to a better embedding rather than allowing the control network to have free rein on the embedding? (3) How do we compare to a state-of-the-art one-shot imitation learning method? (4) How is our performance affected as we go from one-shot to many-shot? (5) Does this method apply to sim-to-real?

We begin by presenting results from two simulated experimental domains that were put forward for MIL~\citep{finn2017one}. We then continue to present results for our own experiment where we perform a placing task using a real-world robot arm, similar to that of MIL's third experimental domain. All 3 experiments are shown in Figure \ref{fig:experiments_summary}. For all experiments, we ensure that our control network follows a similar architecture to MIL~\citep{finn2017one} in order to allow fair comparison. All networks are trained using the ADAM~\citep{kingma2014adam} optimiser with a learning rate of $5\times10^{-4}$, and a batch-size of 64. Further network architecture details are defined in Appendix \ref{app:experimental_details}. Our approach only uses visual information for the demonstrations whilst MIL reports results where the input demonstrations are given with and without actions and robot arm state. For completeness, we have reported all of MIL's results, but our aim is to compare against the results where only visual information is used for input demonstrations. Qualitative results for our approach can be seen in the video\footnote{\label{foot:video}\url{https://sites.google.com/view/task-embedded-control}}.

\subsection{Simulated Reaching}
\label{sec:sim_reach}

\begin{wrapfigure}[17]{r}{7cm}
\centering
\includegraphics[width=1.0\linewidth]{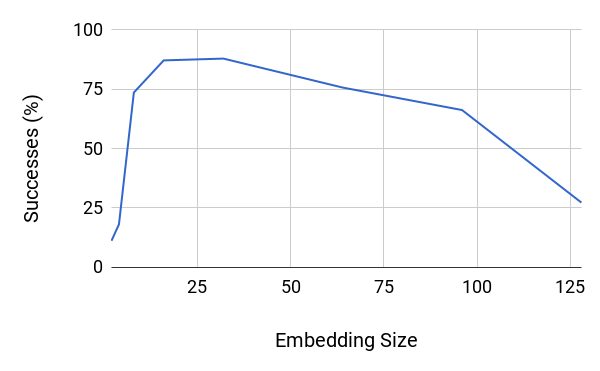}
\caption{How the percentage of success changes as the size of the embedding varies for the simulated reaching domain.}
\label{fig:vary_embedding_size}
\end{wrapfigure} 

The aim of this first experimental domain is to reach a target of a particular colour in the presence of two distractors with different colours. Input to the control network consist of the (current) arm joint angles, end-effector position, and the $80 \times 64$ RGB image, whilst the task-embedding network receives only images (first and last). For details regarding data collection, we point the reader to the Appendix of~\citep{finn2017one}.  Our results (presented in Table \ref{table:sim_results}) show that we outperform MIL by a large margin, as well as other variations of our approach. The results show that the embedding loss is vital for the high success rate, with the exclusion leading to a drop in success of over 70\%. In addition to the embedding loss, the inclusion of the support loss heavily assists the network in learning the task. Note that it is possible to achieve ~33\% on this task by randomly choosing one target to reach for. We believe this is an important note, as it appears that MIL is not capable of learning from one visual demonstration alone on this task, resulting in a policy that randomly selects a colour to reach for. As with MIL, only 1-shot capabilities were explored for this domain.

We also use this experimental domain to see how the embedding size effects the performance, and we show the results in Figure \ref{fig:vary_embedding_size}. By increasing the embedding size, we are increasing the dimensionality of our vector space, allowing a greater number of tasks to be learned. But as Figure \ref{fig:vary_embedding_size} shows, increasing the dimensionality can lead to poor performance. We hypothesise that increasing the embedding size too much can lead to a trivial embedding that does not look for similarities and will thus generalise poorly when encountering new tasks. A balance must be struck between the capacity of the embedding and the risk of overfitting. Although this is an extra hyperparameter to optimise for, Figure \ref{fig:vary_embedding_size} encouragingly suggest that this can take on a wide range of values.

\subsection{Simulated Pushing}
\label{sec:sim_push}

The second experimental domain from~\citep{finn2017one} involves a simulated robot in a 3D environment, where the action space is 7-DoF torque control. The goal is to push a randomly positioned object to the red target in the presence of another randomly positioned distractor, where the objects have a range of shapes, sizes, textures, frictions, and masses. The control network input consists of a $125 \times 125$ RGB image and the robot joint angles, joint velocities, and end-effector pose, whilst the task-embedding network again receives images only. For details regarding data collection, we point the reader to the Appendix of~\citep{finn2017one}. In both the 1-shot and 5-shot case, our method surpasses MIL when using its few-shot ability on visual data alone. Unlike the previous experiment, excluding the support loss is less detrimental and leads to better results than MIL in both the 1-shot and 5-shot case. 

\begin{table}[]
\begin{minipage}{.5\linewidth}
\centering
\begin{tabular}{l | l | c}
                              & Method                   & Success (\%)          \\
\hline \hline
\multirow{7}{*}{\rot{1-Shot}} & MIL (vision+state+action)   & 93.00             \\
                              & MIL (vision)                & \textbf{29.36}*             \\
                              & \textbf{Ours} (vision)                          & \textbf{86.31}        \\
                              & Ours ($\lambda_{ctr}^{\support}=0$)             & 25.68                  \\
                              & Ours ($\lambda_{emb}=0$)                        & 10.48                  \\
                              & Ours ($\sentence=\vec{0}$)                      & 20.30                  \\
                              & Ours (contextual)                               & 19.17                  \\
\end{tabular}
\subcaption{Simulated Reaching Results}
\end{minipage}
\hspace{0.1cm}
\begin{minipage}{.5\linewidth}
\centering
\begin{tabular}{l | l | c}
                              & Method                   & Success (\%)          \\
\hline \hline
\multirow{7}{*}{\rot{1-Shot}} & MIL (vision+state+action)       & 85.81                 \\
                              & MIL (vision+state)              & 72.52                 \\
                              & MIL (vision)                    & \textbf{66.44}        \\
                              & \textbf{Ours} (vision)                              & \textbf{77.25}        \\
                              & Ours ($\lambda_{ctr}^{\support}=0$)                 & 70.72                  \\
                              & Ours ($\lambda_{emb}=0$)                            & 58.56                  \\
                              & Ours ($\sentence=\vec{0}$)                          & 02.49                  \\
                              & Ours (contextual)                                   & 37.61                  \\
\hline
\multirow{8}{*}{\rot{5-Shot}} & MIL (vision+state+action)       & 88.75                 \\
                              & MIL (vision+state)              & 78.15                 \\
                              & MIL (vision)                    & \textbf{70.50}        \\
                              & \textbf{Ours} (vision)                              & \textbf{80.86}         \\
                              & Ours ($\lambda_{ctr}^{\support}=0$)                 & 72.07                  \\
                              & Ours($\lambda_{emb}=0$)                             & 67.12                  \\
                              & Ours ($\sentence=\vec{0}$)                          & -                  \\
                              & Ours (contextual)                                   & -
\end{tabular}
\subcaption{Simulated Pushing Results}
\end{minipage}
\caption{The result for both simulated reaching (a) and simulated pushing (b), for both our full solution, and a series of ablations. In the tables, $\lambda_{ctr}^{\support}=0$ refers to excluding the support loss, $\lambda_{emb}=0$ refers to excluding the embedding loss, $\sentence=\vec{0}$ refers to ignoring the output of the embedding, and instead passing in a zero sentence, and `\textit{contextual}' refers to ignoring the output of the embedding, and passing in one of the support example's images directly to the control network. There is no entry for the final 2 rows of \textit{Table (b)} as these are equivalent to their 1-shot counterpart. *Note that in \textit{Table (a)}, the results reported here for MIL were not reported in the paper, and so the results here are reported from running their publicly available code.}
\label{table:sim_results}
\end{table}

\subsection{Real-world Placing via Sim-to-Real}
\label{sec:simtoreal_place}

The final experiment tests how well our method works when applied to a real robot. Not only that, but we also look at the potential of our method to be used in a sim-to-real context; where the goal is to learn policies within simulation and then transfer these to the real world with little or no additional training (we focus on the latter). This is an attractive idea, as data collection in the real world is often cumbersome and time consuming.

\begin{wrapfigure}[14]{r}{0.4\textwidth}
\centering
\includegraphics[width=0.9\linewidth]{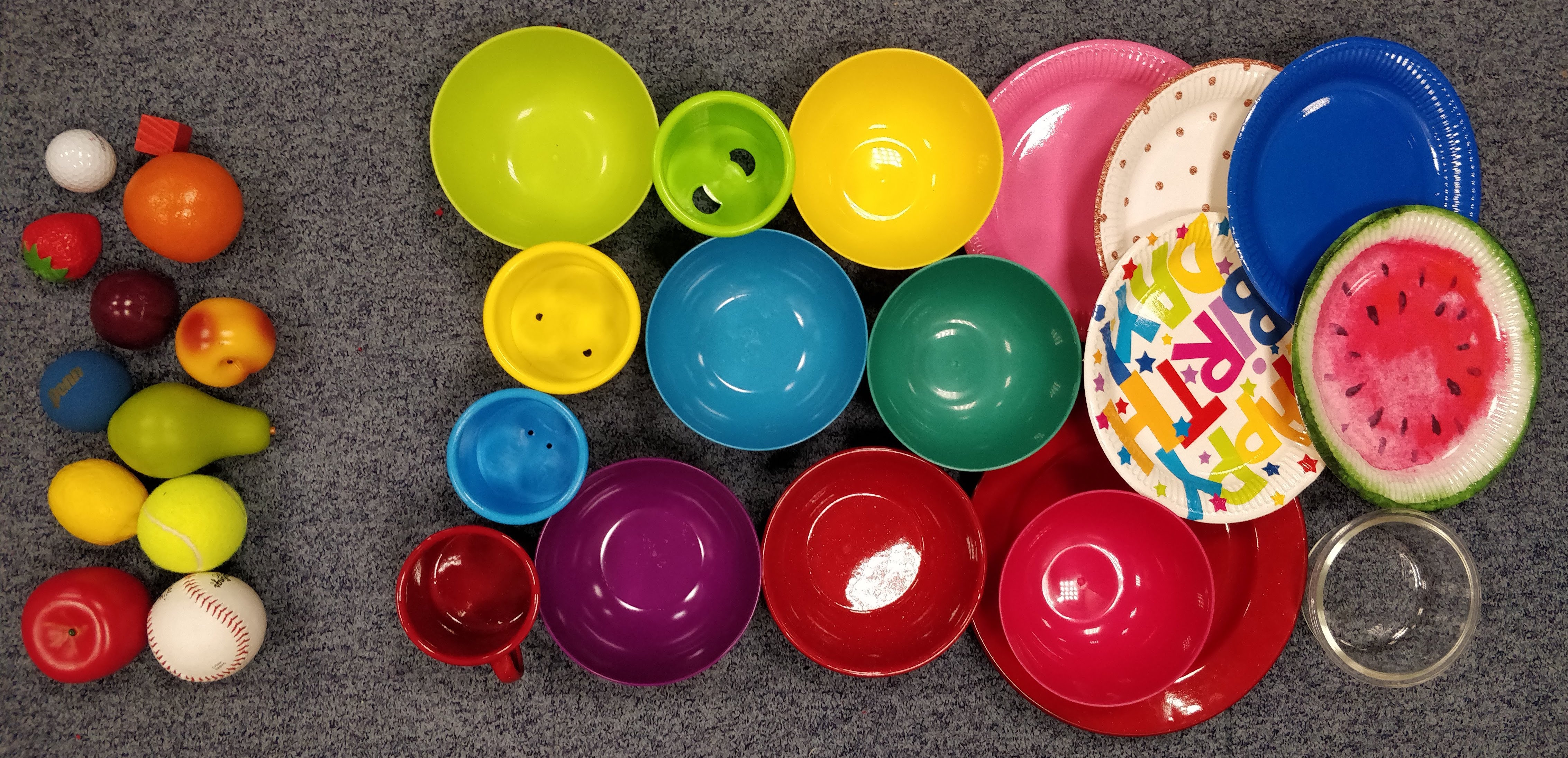}
\caption{The real world test set for the placing domain. Holding objects on the left and placing objects (consisting of bowls, plates, cups, and pots) on the right.}
\label{fig:test_objects}
\end{wrapfigure} 

We run a robotic placing experiment much like the one proposed in MIL, where a robot arm is given the task of placing a held object into a specified container whilst avoiding 2 distractors. The key difference is that our data is collected in simulation rather than real world. As summarised in Figure \ref{fig:front_summary}, our \tecnet{} is trained in simulation with a dataset of 1000 tasks with 12 examples per task. Our containers consist of a selection of 178 bowls from the ShapeNet database~\citep{chang2015shapenet}. To enable transfer, we use domain randomisation; a method that is increasingly being used transfer learned polices from simulation to the real world~\citep{james2017transferring, tobin2017domain, matas2018sim}. We record RGB images of size $160 \times 140$ from an external camera positioned above the robot arm, joint angles, and velocities along a planned linear path for each example. During domain randomisation, we vary the lighting location, camera position, table texture, target object textures and target object sizes, and create procedurally generated holding objects. An example of the randomisation can be seen in Figure \ref{fig:front_summary}.

Once the network has been trained, we randomly select one holding object and 3 placing targets from our test set of real-world objects (shown in Figure \ref{fig:test_objects}); these objects have not been seen before in either simulation or real world. The robot is shown a single demonstration via human teleoperation using the HTC Vive controller. During demo collection, only RGB images and joint angles are collected. A trial is successful if the held object lands in or on the target container after the gripper ihas opened.

One-shot success rates in the real-world is \textbf{72.97\%}, and is based on 18 tasks with 4 examples each (72 trials total), showing that we are able to successfully cross the reality-gap and perform one-shot imitation learning. The majority of our failure cases appeared when the target objects were cups or plates, rather than bowls. We imagine this is due to the fact that our training set only consisted of bowls. Results for the real world evaluation can be seen in the video\footnote{\label{foot:video}\url{https://sites.google.com/view/task-embedded-control}}, and a visualisation of the learnt embedding can be seen in Appendix \ref{app:embedding_vis}.

%===============================================================================

\section{Conclusion}
\label{sec:conclusion}

We have presented \tecnet s, a powerful few-shot learning approach for end-to-end few-shot imitation learning. The method works by learning a compact description of a task via an embedding network, that can be used to condition a control network to predict action for a different example of the same task. Our approach is able to surpass the performance of MIL~\citep{finn2017one} for few-shot imitation learning in two experimental domains when only visual information is available. Unlike many other meta-learning approaches, our method is capable of continually learning new tasks without forgetting old ones, and without losing its few-shot ability. Moreover, we demonstrate that the few-shot ability can be trained in simulation and then deployed in the real world. Once deployed, the robot can continue to learn new tasks from single or multiple demonstrations. 

Similar to other meta-learning approaches, we expect the approach to perform poorly when the new task to learn is drastically different from the training domain; for example, a \tecnet{} trained to place items in containers would not be expected to learn few-shot pushing. Having said that, if the training set were to include a wide range of tasks, generalising to a broad range of tasks may be possible, and so this is something that should be looked at further. Parallel work has shown extensions to MIL that infer policies from human demonstration~\citep{yu2018one}. As our method inherently only uses visual information, we are also keen to investigate the inclusion of human demonstrations to our approach.

%===============================================================================

% The maximum paper length is 8 pages excluding references and acknowledgements, and 10 pages including references and acknowledgements

\clearpage
\acknowledgments{Research presented in this paper has been supported by Dyson Technology Ltd. We thank the
reviewers for their valuable feedback.}

%===============================================================================

\bibliography{corl}  % .bib

\newpage
\appendix

\section{Experimental Details}
\label{app:experimental_details}

In this section we provide additional experiment details, network architecture, and hyperparameters. In all cases the task-embedding network and control network use a convolutional neural network (CNN),  where each layer is followed by layer normalisation~\citep{ba2016layer} and an $elu$ activation function~\citep{clevert2015fast}, except for the final layer, where the output is linear for both the task-embedding and control network. Optimisation was performed with Adam~\citep{kingma2014adam} with a learning rate of $5 \times 10^{-4}$, and lambdas were set as follows: $\lambda_{emb} = 1.0$, $\lambda_{ctr}^{\support} = 0.1$, $\lambda_{ctr}^{\query} = 0.1$.

\subsection{Simulated Reaching}

The CNN consists of $3$ strided convolution layers, each with $40$ ($3\times3$) filters, followed by $4$ fully-connected layers consisting of $200$ neurons. Input consists of a $80 \times 64$ RGB image and the robot proprioceptive data, including the arm joint angles and the end-effector position. The proprioceptive is concatenated to the features extracted from the CNN layers of the control network, before being sent through the fully-connected layers. The output of the embedding network is a vector of length $20$. The output corresponds to torques applied to the two joints of the arm. The task is considered a success if the end-effector comes within 0.05 meters of the goal within the last 10 timesteps. Further information regarding this task can be accessed from \citet{finn2017one}.

\subsection{Simulated Pushing}

The CNN consists of $4$ strided convolution layers, each with $16$ ($5\times5$) filters, followed by $3$ fully-connected layers consisting of $200$ neurons. Input consists of a $125 \times 125$ RGB image and the robot proprioceptive data, including the joint angles, joint velocities, and end-effector pose. The proprioceptive is concatenated to the features extracted from the CNN layers of the control network, before being sent through the fully-connected layers. The output of the embedding network is a vector of length $20$. The output of the control network corresponds to torques applied to the $7$ joints of the arm. The task is considered a success if the robot pushes the centre of the target object into the red target circle for at least 10 timesteps within 100-timestep episode. Further information regarding this task can be accessed from \citet{finn2017one}.

\subsection{Real-world Placing}

The CNN consists of $4$ strided convolution layers, each with $16$ ($5\times5$) filters, followed by $4$ fully-connected layers consisting of $100$ neurons. Input consists of a $125 \times 125$ RGB image and the robot proprioceptive data, including just the joint angles. The proprioceptive is concatenated to the features extracted from the CNN layers of the control network, before being sent through the fully-connected layers. The output of the embedding network is a vector of length $20$. The output of the control network corresponds to torques applied to the $7$ joints of a Kinova Mico 7-DoF arm. There is also an additional auxiliary end-effector position output that is learned via an $L2$ distance between the prediction and the ground truth during simulation training. The task is considered a success if the robot drops the held object into the correct target container.

As a note, we also experimented with using a U-Net architecture~\citep{ronneberger2015u} for the control network, where the sentence is concatenated to the image features at the bottleneck, but our experiments showed that channel-wise concatenation at the input layer of the control network worked just as well.

\section{Sim-to-Real Embedding Visualisation}
\label{app:embedding_vis}

In this section we show t-SNE~\citep{maaten2008visualizing} visualisation of the learnt embedding of the real-world placing task of Section \ref{sec:simtoreal_place}. Note that the \tecnet{} was trained entirely in simulation without having seen any real-world data. In order to visualise how the embedding looks on real-world data, we collect a dataset of 164 tasks, each consisting of 5 demonstrations. Each demonstration consists of a series of RGB images that were collected via human teleoperation using the HTC Vive controller. Each demonstration in these visualisations are represented via the final frame of that demonstration.

\begin{figure}
\centering
\includegraphics[width=1.0\linewidth]{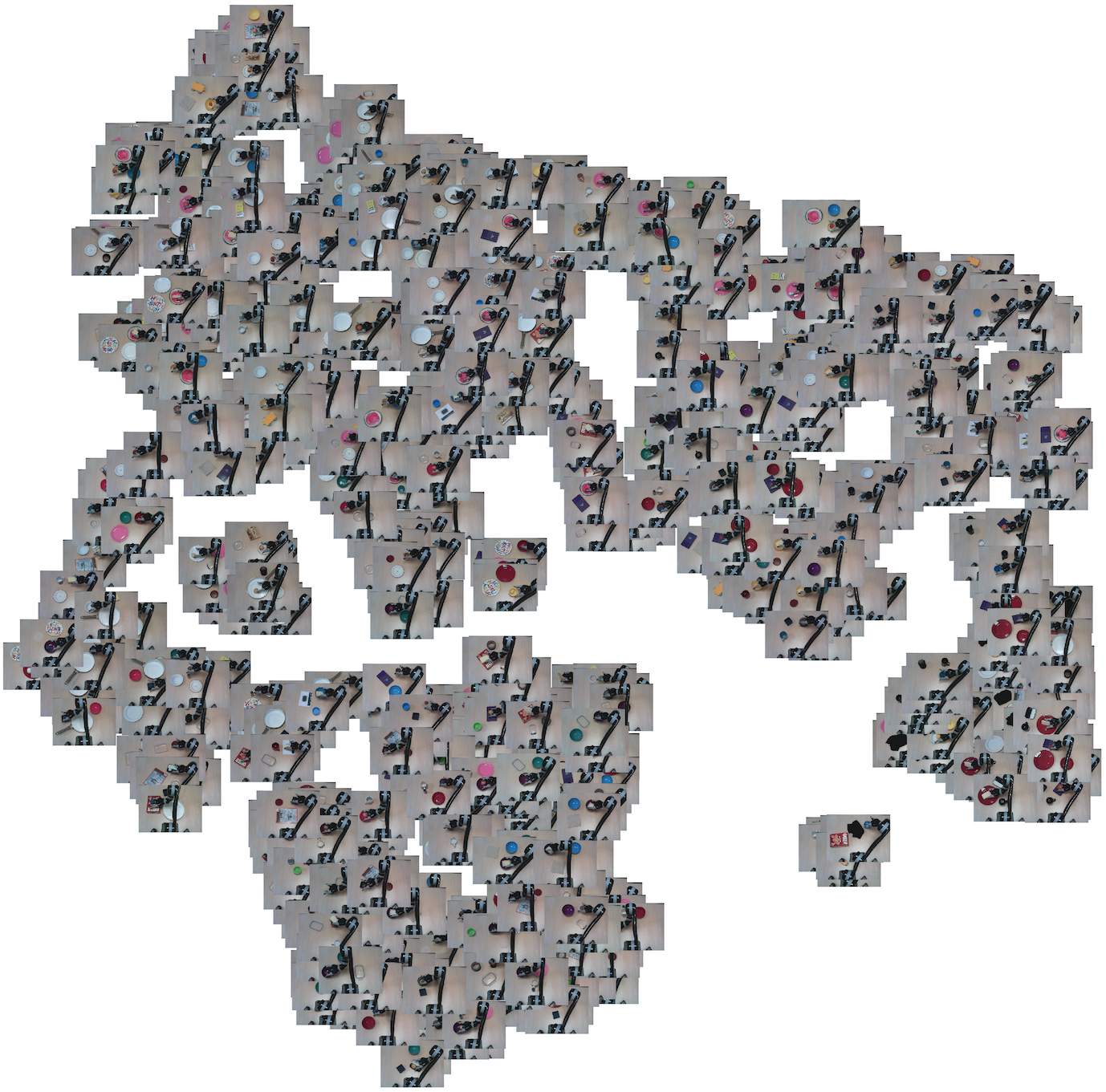}
\caption{A t-SNE visualisation of the individual sentences of each of the demonstrations learnt by the task-embedding network. We embed 5 demonstrations (without averaging) across each of the 164 tasks. The aim of the visualisation is to illustrate how examples of the same task relate with each other. The result shows that the task-embedding network does indeed learn to place examples of the same tasks next to each other, whilst also placing other, visually similar, tasks nearby.}
\label{fig:simtoreal_tsne_embeddings}
\end{figure}

\begin{figure}
\centering
\includegraphics[width=1.0\linewidth]{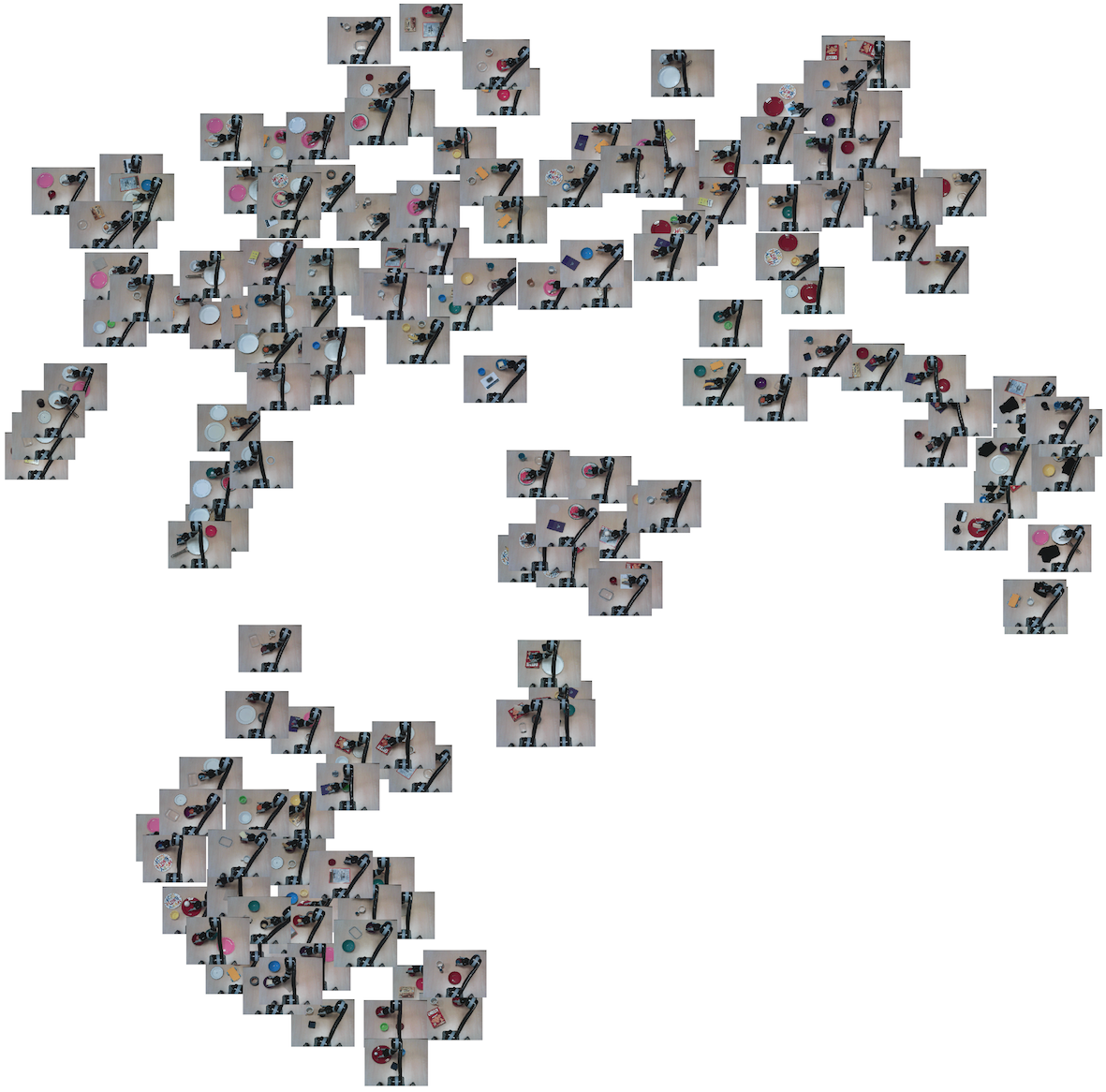}
\caption{A t-SNE visualisation of the combined sentences learnt by the task-embedding network. We embed 5 demonstrations and average to get the task sentence for each of the 164 tasks. Given that we are only plotting the combined sentences, this can be seen as a more legible version of Figure \ref{fig:simtoreal_tsne_embeddings}, focusing on how tasks relate to other tasks, rather than how examples of the same task relate with each other.}
\label{fig:simtoreal_tsne_sentences}
\end{figure}

\end{document}